\documentclass[letterpaper, 10 pt, conference]{ieeeconf}  

\IEEEoverridecommandlockouts                              

\usepackage{graphics} 
\usepackage{graphicx,import, multicol}
\usepackage{subfigure}
\usepackage{algorithmic}
\usepackage{amsmath,amssymb,amsfonts}

\usepackage{tikz}                      
\usepackage{tikzscale}
\usepackage{relsize}
\usetikzlibrary{calc, backgrounds, shapes, arrows, shapes}
\usepackage{pgfplots}                  
\pgfplotsset{compat=newest} 
\usetikzlibrary{plotmarks}

\newcommand\copyrighttext{%
	\footnotesize \copyright~2018 IEEE. Personal use of this material is permitted. Permission from IEEE must be obtained for all other uses, in any current or future media, including reprinting/republishing this material for advertising or promotional purposes, creating new collective works, for resale or redistribution to servers or lists, or reuse of any copyrighted component of this work in other works.  DOI: 10.1109/ITSC.2018.8569235}%
\newcommand\copyrightnotice{%
	\begin{tikzpicture}[remember picture,overlay]
	\node[anchor=south,yshift=10pt] at (current page.south) {\fbox{\parbox{\dimexpr\textwidth-\fboxsep-\fboxrule\relax}{\copyrighttext}}};
	\end{tikzpicture}%
}

\title{\LARGE \bf
Environment Perception Framework Fusing Multi-Object Tracking, Dynamic Occupancy Grid Maps and Digital Maps
}

\author{Fabian Gies, Andreas Danzer and Klaus Dietmayer
\thanks{F. Gies, A. Danzer, and K. Dietmayer are with the Institute of Measurement, Control and Microtechnology, Ulm University, 89081 Ulm, Germany \{\tt\small fabian.gies, andreas.danzer, klaus.dietmayer\}@uni-ulm.de}%
}

\renewcommand{\vec}[1]{{\mathrm{#1}}}
\newcommand{\vecset}[1]{\boldsymbol{\vec{#1}}} 
\newcommand{\trans}{^\text{\rm \textup{T}}}

\DeclareMathOperator{\atantwo}{arctan2}
\newcommand{\zweinorm}[1]{\ensuremath \left\| #1 \right\|_2}

\newcommand{\figurename}{Figure~}

\graphicspath{{img/}}

\newlength\figH
\newlength\figW
\begin{document}

\maketitle
\copyrightnotice
\thispagestyle{empty}
\pagestyle{empty}

\begin{abstract}
    Autonomously driving vehicles require a complete and robust perception of the local environment. A main challenge is to perceive any other road users, where multi-object tracking or occupancy grid maps are commonly used. The presented approach combines both methods to compensate false positives and receive a complementary environment perception. Therefore, an environment perception framework is introduced that defines a common representation, extracts objects from a dynamic occupancy grid map and fuses them with tracks of a Labeled Multi-Bernoulli filter. Finally, a confidence value is developed, that validates  object estimates using different constraints regarding physical possibilities, method specific characteristics and contextual information from a digital map. Experimental results with real world data highlight the robustness and significance of the presented fusing approach, utilizing the confidence value in rural and urban scenarios.
\end{abstract}


%
\IEEEpeerreviewmaketitle

\section{Introduction}
\label{sec:introduction}
A major challenge for autonomously driving vehicles is a complete and robust perception of the surroundings. For this reason, the vehicle is equipped with a large variety of sensors that generate a dense and precise depiction of the local environment. The main objective of these sensor measurements is to detect and track any other road users.
Algorithms using temporal filtering of the sensor measurements are commonly used and are a well studied topic \cite{BarShalom1988}. These object tracking approaches apply \textit{object-model-based} prior information and use Bayesian filtering techniques to suppress uncertainties and clutter. Further, multi-object tracking approaches \cite{Mahler2007} are able to detect and estimate the state of multiple objects at once. Here, measurements and tracks have to be associated and assigned. To overcome explicit associations, the usage of a Random Finite Set (RFS) for probabilistic estimations of the objects states and the sets cardinality, showed impressive results \cite{Mahler2007,Reuter2014b}. Despite the great success, multi-object tracking in urban scenarios with a large variety and amount of traffic participants is still a tough challenge.
A second approach for an environment perception are occupancy grid maps \cite{Elfes1989,Thrun2005}. For this reason, the local environment is separated in single independent grid cells, where the occupancy probability of each cell is estimated using the sensor measurements. Due to the \textit{object-model-free} representation detecting any object type, e.g. pedestrians, cars and trucks, is possible. The Dynamic Occupancy Grid Map (DOGMa) \cite{Nuss2017,Tanzmeister2016} is an extension to the classical grid map and is able to estimate velocities for dynamic areas. These grid map implementations showed significant results for a generic environment perception. However, a huge disadvantage for an object detection is a missing association of grid cells to the corresponding object. Steyer et al. \cite{Steyer2017} recently published promising results to combine a grid map and a multi-object tracking approach. Here, object hypotheses are generated by clustering cells of an evidential DOGMa and tracked with a subsequent unscented Kalman filter. Additionally, Asvadi et al. \cite{Asvadi2015} presented a similar approach using a 2.5D motion grid. Furthermore, different approaches \cite{Jungnickel2014,Yuan2017} showed the effectiveness of density-based clustering, e.g. Density-Based Spatial Clustering of Applications with Noise (DBSCAN) to cluster cells as object hypotheses from an occupancy grid map. 

Certainly, fully automated vehicles depend on a accurate estimation of the objects and their states. Especially, planning and control modules, like \cite{Ulbrich2017} presented, malfunction of an error propagation of incorrect data. In the presented approach an environment perception framework is introduced, where the advantages of a multi-object tracking and a DOGMa are combined to receive a complete and robust estimation of other road users. 
A common representation of any object state is defined using an ego-stationary coordinate system \cite{Weiss2011}. Since, the work of \cite{Gindele2009, Hosseinyalamdary2015} and \cite{Nuss2014} showed the significance of contextual information, a digital map with roads and buildings is incorporated into the framework. Additionally, a DOGMa object extraction is presented using a clustering of cells to estimate a full object state. Finally, extracted tracks of a Labeled Multi-Bernoulli (LMB) filter \cite{Reuter2014} and the extracted DOGMa objects are transmitted to a high-level fusion module. Here, the object estimates are validated to ensure a complete and accurate object list. For that reason, a confidence value is developed that determines the quality of an object estimate regarding \textit{physical, module} and \textit{digital~map} constraints.

The remaining paper is organized as follows: Section \ref{sec:background} reviews the basic properties of an LMB filter and the DOGMa. In Section \ref{sec:framework} the presented environment perception framework including all functional modules is described. Subsequently, in Section \ref{sec:highlevelfusion} the fusion of the object estimates considering a confidence value is explained, followed in Section \ref{sec:experiments} by experimental results with real world data. This work is closed in Section \ref{sec:conclusion} with a summary and outlook.

\section{Background}
\label{sec:background}
This section gives a short summary of the characteristics and parameters of a multi-object tracking using the Labeled Multi-Bernoulli (LMB) filter. Furthermore, the occupancy grid map and especially its extension, the Dynamic Occupancy Grid Map (DOGMa) is reviewed. 

\subsection{Labeled Multi-Bernoulli Filter}
\label{sec:lmbfilter}
The LMB filter \cite{Reuter2014} is an accurate and fast multi-object tracking filter using Random Finite Sets (RFSs) \cite{Mahler2007}. The multi-object state is represented by the RFS $\text{X} = \{\mathrm{x}^{(1)},\dots,\mathrm{x}^{(N)}\} \subset \mathbb{X}$, where $\mathrm{x}^{(i)} \in \mathbb{X}$ are the single-target state vectors and $\mathbb{X}$ the state space. If not empty, a multi-Bernoulli RFS comprises multiple independent Bernoulli RFSs that represent the spatial distribution $p(\mathrm{x})$ with a probability $r$. In a scenario with multiple objects, a major challenge is to estimate the current state as well as the identity of an object. Therefore, Vo et al. \cite{Vo2013} introduced the class of labeled RFSs. Here, a distinct label $\ell \in \mathbb{L}$ is appended to each state vector $\mathrm{x} \in \mathbb{X}$, where $\mathbb{L}$ is a finite label space. The multi-object posterior LMB RFS is represented with a parameter set ${\boldsymbol \pi = \{(r^{(\ell)},p^{(\ell)}(\mathrm{x}))\}_{\ell \in \mathbb{L}}}$. So, any tracked object comprise an unique label $\ell$, an existence probability $r^{(\ell)}$ and a spatial distribution $p^{(\ell)}(\mathrm{x})$.
An advantage of the LMB filter is modeling uncertainty in data association implicitly. Considering that, the spatial distribution of a track comprises the association of tracks to multiple measurements. For a detailed explanation and the equations of an LMB filter, see \cite{Reuter2014b}.

\subsection{Occupancy Grid Mapping}
\label{sec:backgroundGM}
An occupancy grid map is a discrete representation of the environment, where the local surroundings of a vehicle is separated into single grid cells~$c$. In classical approaches \cite{Elfes1989,Thrun2005} an occupancy probability of each cell is estimated using sensor measurements, that are updated with the Bayesian rule. For modeling inconsistencies of free and occupied cells the Dempster-Shafer theory of evidence \cite{Shafer1976} is implemented, where each cell holds a mass for free $M_F \in (0,1)$ and occupied $M_O \in (0,1)$. The Bayesian occupancy probability can be calculated with $ p_c(O) = 0.5 \cdot M_O + 0.5 \cdot \left( 1.0 - M_F \right)  \in (0,1)$.
These classical grid maps assume, that the environment contain only static objects and each grid cell is independent of all other grid cell states. 

An extension of the classic occupancy grid map is the DOGMa, that additionally enables an estimation of the dynamic environment \cite{Tanzmeister2016,Nuss2017}. In the presented approach, a particle filter estimates a velocity in $x$ and $y$ direction of each occupied cell, what corresponds to the implementation of \cite{Nuss2017}. The DOGMa provides cells in $\mathbb{R}^{W \times H}$ with width~$W$ and height~$H$ around the vehicle pointing to stationary coordinates $x$ and $y$, respectively. The spatial resolution of grid maps depend on the quadratic grid cell size $a_c$. The higher the resolution the more precisely an object can be detected, given a sensors uncertainty. Finally, each grid cell holds a state 
\begin{equation}
	\vec{s}_c = \left\{M_O, M_F, \vec{p}, \vec{v}, \vec{P} \right\},
	\label{eq:cellstate}
\end{equation}
with a mass for occupied $M_O$ and free space $M_F$. Further, a two dimensional cell position $\vec{p} = [x,y]\trans$, with velocity $\vec{v} = [v_x, v_y]\trans $ and the corresponding covariance matrix
\begin{equation}
	\vec{P} =	
	\begin{bmatrix}
	\sigma^2_{v_x v_x} 	    & \sigma^2_{v_x v_y} \\
	\sigma^2_{v_y v_x} & \sigma^2_{v_y v_y}
	\end{bmatrix}
\end{equation}
is given. For a more detailed mathematical description of the DOGMa, refer to \cite{Nuss2017}.

\section{Environment Perception}
\label{sec:framework}
In this section, the system architecture with all necessary components is presented. Here, all functional modules are described with their main characteristics and attributes. 

\subsection{System Architecture}
\begin{figure}[!t]
	\vspace{2mm}
	\centering
	\resizebox{0.8\columnwidth}{!}{\includegraphics{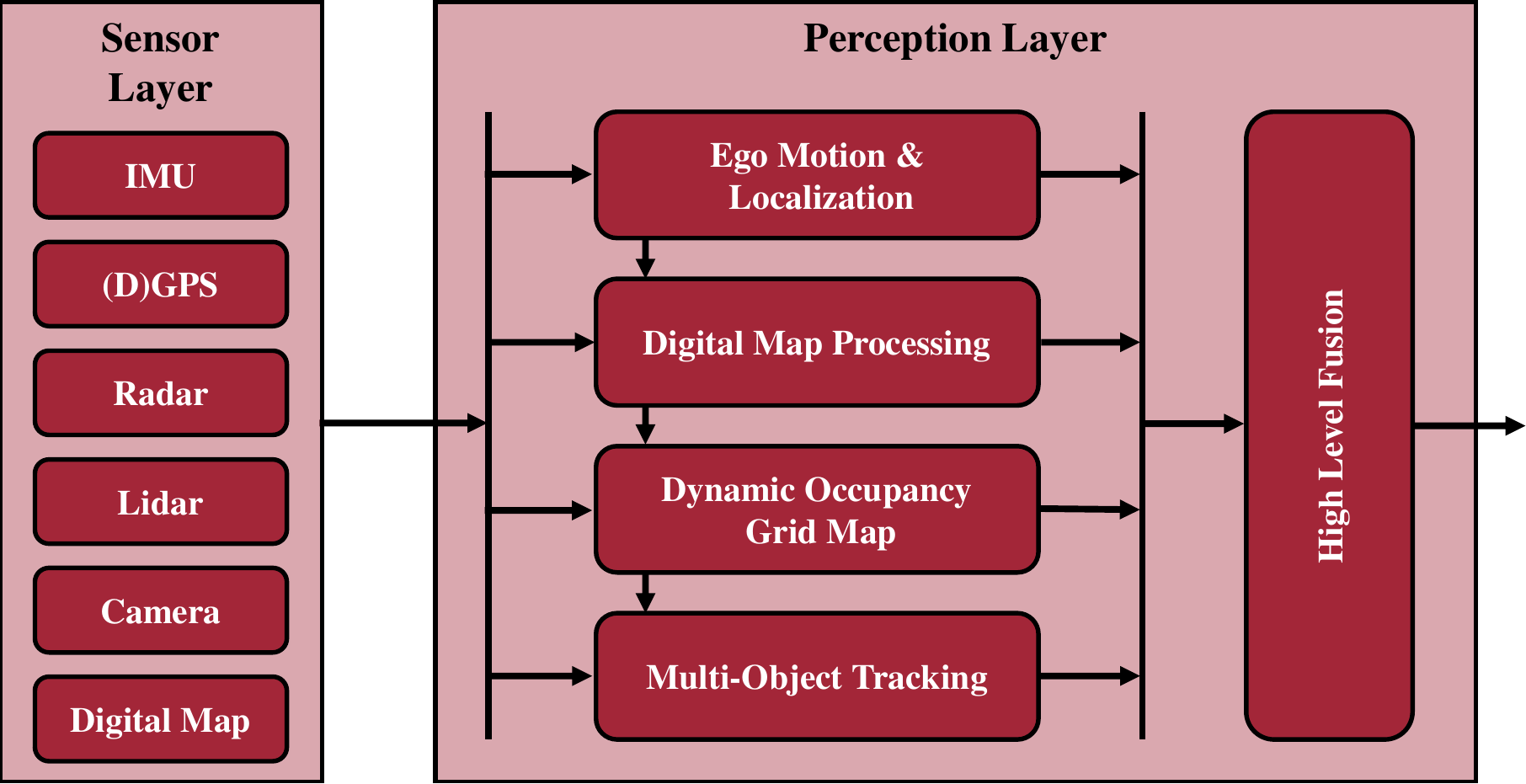}}
	\caption{System architecture of the environment perception.}
	\label{fig:framework}
\end{figure}

The environment perception framework is shown in \figurename\ref{fig:framework}. The system architecture comprises out of a sensor layer with multiple sensor types and a perception layer with various functional modules resulting in a high-level fusion module. In the sensor layer a GPS timestamp is applied to every measurement, that will be used to queue and sort input data. Primarily, the presented perception layer focuses on perceiving other road users like vehicles, cyclist, pedestrians or trucks. For this reason, a module for estimating the ego motion, a multi-object tracking and a DOGMa is implemented. The results of these functional modules are passed to a high-level fusion module. A main idea is the generic structure of the functional modules, where each can be replaced with any other methods generating equal, worse or even better results. In the following, the functional modules with their input and output interfaces are described.

\subsection{Ego Motion and Localization}
As a common representation for any external objects, an ego-stationary coordinate system is defined. For that purpose, the measurements of an inertial measurement unit (IMU) sensor are applied to a dead reckoning motion model \cite{Weiss2011}. The ego vehicle's state
\begin{equation}
	\vec{s}_{ec} = [x,y,v,a,\varphi,\omega]\trans
\end{equation}
is given with a two dimensional position $x,y$, velocity $v$, acceleration $a$, orientation $\varphi$ and a turn rate $\omega$. Starting from the initial state, where $x, y \text{ and } \varphi$ are equal zero, the IMU measurements are used to calculate the vehicle's motion with a constant turn rate and acceleration (CTRA) model \cite{Schubert2008}. Since, the measurements of the IMU can be noisy, a standard linear kalman filter estimates the dynamic states $v, a \text{ and } \omega$. 
Additionally, to utilize information from a global digital map a localization of the ego vehicle in the world is necessary. Accordingly, the ego vehicle's global state 
\begin{equation}
	\vec{s}_{gc} = [\text{UTM}_E,\text{UTM}_N,v,a,\varphi_{gc},\omega]\trans 
\end{equation} 
is estimated using a differential global positioning system (DGPS) and represented in Universal Transverse Mercator (UTM) coordinates. Thus, each point ${\vec{p}_{gc} = [\text{UTM}_E,\text{UTM}_N,\varphi_{gc}]\trans} $ in global coordinates can be transformed to the corresponding ego-stationary coordinates $\vec{p}_{ec} = [x,y,\varphi]\trans$ and vice versa. The ego vehicle's state is directly transmitted to the remaining modules.

\subsection{Digital Map Processing}
\label{seq:digitalmap}
The digital map contains road courses and buildings from two different sources. Firstly, the Openstreetmap (OSM) \cite{OSM2018} is used, where especially the contour of buildings is a valuable information. Even though, OSM contains inaccurate information it can be used as prior. Here, the global corner points of OSM buildings $\vecset{B}$ are transformed to the ego-stationary coordinates. 

Additionally, a set of lanes $\vecset{L}$, were recorded with the experimental vehicle of Ulm University \cite{Kunz2015} using the a high precise DGPS. Each lane represents the reference line of a single road using a vector of two dimensional equidistant points $\vec{p}_{gc}$ in global coordinates and an unique line ID. Further, to employ the lanes in an appropriate way, an iterative end point-fit algorithm \cite{Ramer1972} approximates the lanes. As a result, every lane is approximated with a set of rectangles $\vecset{R}$ each consisting of a center point with orientation, width, length and an identifier. The number of required rectangles and their corresponding length results from the algorithm and depend on the road curvature. Since, the rectangles depict parts of a lane, the calculated orientation is interpreted as the course of a road. The width is specified with a fixed value for the offline map, but could be calculated using an online road detection. A benefit of the rectangles is an easy association to any road user. Finally, the lanes and rectangles can be incorporated to the ego-stationary coordinates.

\begin{figure}[!t]
	\vspace{2mm}
	\centering
	\setlength{\figH}{4cm}
	\setlength{\figW}{\columnwidth}
	\resizebox{0.8\columnwidth}{!}{
\small 
\input{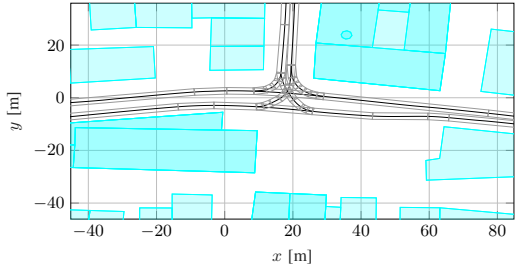}}
	\vspace{-3mm}
	\caption{Section of the digital map in Ulm inner city with OSM buildings (cyan), recorded lanes (black) and approximated rectangles (grey).}
	\label{fig:digitalmap}
\end{figure}

The complete digital map $\mathcal{M} = \left\{ \vecset{B}, \vecset{L}, \vecset{R} \right\}$ is shown in \figurename{\ref{fig:digitalmap}}, where the OSM buildings (cyan), the lanes (black) and the rectangles (grey) are visualized in an urban area. 
%
\subsection{Multi-Object Tracking}
An \textit{object-model-based} tracking of other road user is an essential task for the environment perception. Consequently, every object should be detected and their state estimated over subsequent time steps. For a temporal filtering of sensor measurements and associating these to the dynamic state of multiple road users at once, a multi-object tracking is used. In the presented perception framework, a generic functional module architecture is realized, so that the multi-object tracking can be replaced with similar implementations.

In this work an LMB filter is implemented to track vehicles using radar and camera detections, and is noted with $\mathcal{T}$. For detailed information of the filter processing, see Section \ref{sec:lmbfilter}. The implemented LMB filter uses a CTRA motion model to predict the tracked objects and consequently, the orientation and velocity of tracks can be estimated directly.

To extract tracks of the posterior LMB distribution the existence probability $r^{(\ell)}$ of a track with label $\ell$ has to exceed a minimum threshold $\vartheta_r$. Finally, the state of extracted tracks 
\begin{equation}
	\hat{\vec{s}}_\mathcal{T} = \{\vec{p}_{rp}, v,a, \varphi, \omega, \vec{B}, \vec{P}, r, c, \ell  \}
\label{eq:trackstate}
\end{equation}
consists of a two dimensional reference point position $\vec{p}_{rp} = [x,y]\trans$ with the set 
\begin{equation}
rp = \big\{  \text{b, bl, l, fl, f, fr, r, br} \big\},
\label{eq:refpoint} 
\end{equation} 
of labels back (b), back left (bl), left (l), front left (fl), front (f), front right (fr), right (r) or back right (br), the absolute velocity $v$, acceleration $a$, orientation $ \varphi $ and a turn rate $\omega$. In addition, the length and width of an object is estimated resulting in an orientated bounding box
\begin{equation}
\vec{B} = \left[  \vec{p}_{\text{{bl}}}, \vec{p}_{\text{{fl}}}, \vec{p}_{\text{{fr}}}, \vec{p}_{\text{{br}}} \right].
\label{eq:bb}
\end{equation}
The uncertainty of the states is estimated with the covariance matrix $\vec{P}$ accordingly. Finally, the existence probability $r$, the class type $c$ and an unique label $\ell$ is stated. After the filter processing cycle is finished, the set $\vecset{S}_\mathcal{T}$ includes all extracted tracks $ \hat{\vec{s}}_\mathcal{T} \in \vecset{S}_\mathcal{T}$ and is transmitted to the high-level fusion module at every sample time $t_\mathcal{T}$. 
\subsection{Dynamic Occupancy Grid Map}
In the presented framework, the DOGMa separates the local environment into single grid cells $c$ and uses lidar measurements to estimate the occupancy, free space and velocity per cell. The implemented DOGMa is further described in Section \ref{sec:backgroundGM} and in the following labeled with $\mathcal{G}$. Here, the essential characteristics and parameters of using a DOGMa in the environment perception framework are highlighted. 

A main advantage of using a DOGMa for the object detection is the \textit{object-model-free} environment perception. Thus, any road users like cars, pedestrian, bicycle or trucks can be detected. To estimate the dynamic areas particles are predicted with a constant velocity (CV) model and updated with the next sample of a measurement grid. In the end, every cell holds a state (\ref{eq:cellstate}), and is updated with sample time~$t_\mathcal{G}$. Since, no turn rate is depicted in a CV model, the cell orientation is assumed constant and calculated with $\varphi_c = \atantwo(v_y, v_x) $. Consequently, objects moving in sharp curves can have an uncertain estimation for the orientation. Furthermore, any cell state of a DOGMa is independent of all other cell states, thus there is no association between cells and the corresponding objects, that caused the measurements. For that reason, in the presented approach a grid map object extraction is developed to assign each cell to an object and finally receive a full state description of any road user perceived with the DOGMa. 
\subsection{Grid Map Object Extraction}
\begin{figure}[!t]
	\vspace{2mm}
	\centering  
	\setlength{\figH}{9cm}
	\setlength{\figW}{\columnwidth} 
	\subfigure[\label{subfig:a:gridmapextraction}]{\includegraphics[width=0.3\columnwidth]{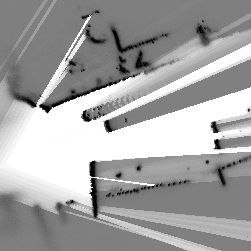}} 
	\subfigure[\label{subfig:b:gridmapextraction}]{\includegraphics[width=0.3\columnwidth]{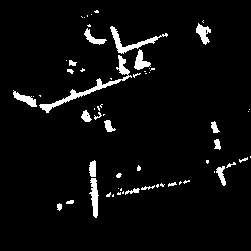}}
	\subfigure[\label{subfig:d:gridmapextraction}]{\includegraphics[width=0.3\columnwidth]{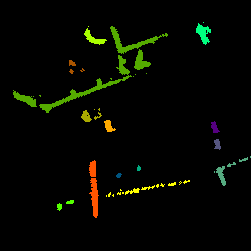}}
	\subfigure[\label{subfig:c:gridmapextraction}]{\includegraphics[width=0.3\columnwidth]{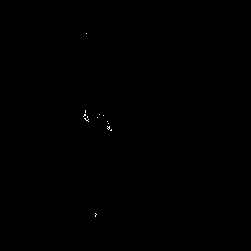}}
	\subfigure[\label{subfig:e:gridmapextraction}]{\includegraphics[width=0.3\columnwidth]{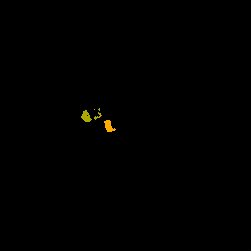}}
	\subfigure[\label{subfig:f:gridmapextraction}]{\resizebox{0.3\columnwidth}{!}{
\small 
%
%
\definecolor{mycolor1}{rgb}{0.66667,0.66667,0.00000}%
\definecolor{mycolor2}{rgb}{1.00000,0.66667,0.00000}%
\begin{tikzpicture}

\begin{axis}[%
width=0.951\figW,
height=\figH,
at={(0\figW,0\figH)},
scale only axis,
xmin=5,
xmax=35,
xlabel style={font=\color{white!15!black}},
xlabel={$x_{vc}$ [m]},
ymin=-14,
ymax=14,
ylabel style={font=\color{white!15!black}},
ylabel={$y_{vc}$ [m]},
axis background/.style={fill=white},
xmajorgrids,
ymajorgrids
]
\addplot [color=mycolor1, line width=1.2pt, draw=none, mark=o, mark size=0.2pt, mark options={solid, mycolor1}, forget plot]
  table[]{img/tikz/objects534-1.tsv};
\addplot [color=mycolor1, line width=1.2pt, draw=none, mark=square,mark size=0.2pt, mark options={solid, mycolor1}, forget plot]
  table[]{img/tikz/objects534-2.tsv};
\addplot [color=mycolor1, line width=1.2pt, forget plot]
  table[]{img/tikz/objects534-3.tsv};
\node[right, align=left]
at (axis cs:18.847,2.873) {4};
\addplot [color=mycolor1, line width=1.2pt, draw=none, mark=x, mark options={solid, mycolor1}, forget plot]
  table[]{img/tikz/objects534-4.tsv};
\addplot [color=mycolor1, line width=1.2pt, forget plot]
  table[]{img/tikz/objects534-5.tsv};
\addplot [color=mycolor2, line width=1.2pt, draw=none, mark=o,  mark size=0.2pt,mark options={solid, mycolor2}, forget plot]
  table[]{img/tikz/objects534-6.tsv};
\addplot [color=mycolor2, line width=1.2pt, draw=none, mark=square, mark size=0.2pt, mark options={solid, mycolor2}, forget plot]
  table[]{img/tikz/objects534-7.tsv};
\addplot [color=mycolor2, line width=1.2pt, forget plot]
  table[]{img/tikz/objects534-8.tsv};
\node[right, align=left]
at (axis cs:21.104,0.377) {7};
\addplot [color=mycolor2, line width=1.2pt, draw=none, mark=x, mark options={solid, mycolor2}, forget plot]
  table[]{img/tikz/objects534-9.tsv};
\addplot [color=mycolor2, line width=1.2pt, forget plot]
  table[]{img/tikz/objects534-10.tsv};
\end{axis}
\end{tikzpicture}%
}}
	\caption{DOGMa sample visualizing the grid map object extraction approach from (a) to (f) with an oncoming and a leading vehicle passing each other in front of the ego vehicle. (a) Occupancy probability of the DOGMa. (b) Search Mask. (c) Cluster Cells. (d) Validation Mask. (e) Validate Clusters. (f) Extracted objects in vehicle coordinates with a bounding box, velocity, orientation and an unique label.}
	\label{fig:gridmapextraction}
\end{figure}
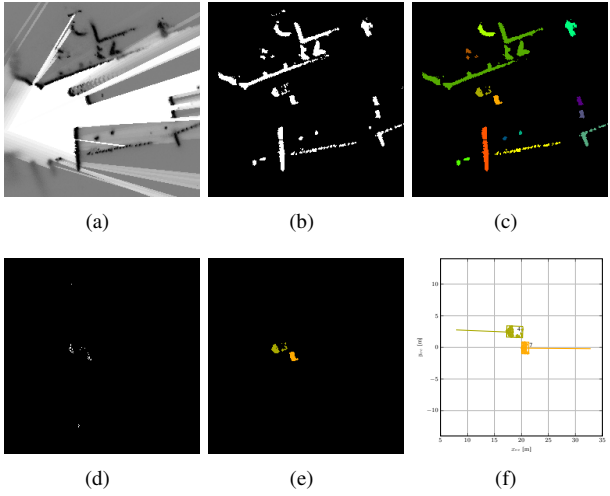

In the following, every single step is described to extract objects from the grid map cell states, what is visualized in \figurename{\ref{fig:gridmapextraction}}. An exemplary scene shows \figurename{\ref{subfig:a:gridmapextraction}}, where the occupancy probability $p_c(O)$ is indicated with dark pixel for occupied cells and white pixels for free space. Unknown cells with $p_c(O) = 0.5$, or $M_F = 0$ and $M_O = 0$ are marked gray. The scene is recorded in a urban area, where two vehicles passing each other in front of the ego vehicle. Hereafter, the set $\vecset{C}_\mathcal{G}$ of all cell states of the entire DOGMa $\mathcal{G}$ is evaluated: 
\begin{equation}
\vec{s}_c = \left\{M_O, M_F, \vec{p}, \vec{v}, \vec{P} \right\} \in \vecset{C}_\mathcal{G}.
\label{eq:cellstategme}
\end{equation} 

\subsubsection{Search Mask}
The search mask is the basis for assigning cells to a corresponding object. Here, the search mask identifies cells, where any measurement was incorporated. The set of cell states in the search mask
\begin{equation}
	\vecset{C}_s = \Big\{ \vec{s}_c \in \vecset{C}_\mathcal{G} | M_O  > \varepsilon_{M_O}  \Big\}
\end{equation} 
includes any cells with a minimum occupancy mass $\varepsilon_{M_O}$. This threshold ensures to remove unknown cells. \figurename{\ref{subfig:b:gridmapextraction}} shows the remaining cells with white pixels.
\subsubsection{Cluster Cells}
In this step a cluster is formed using the DBSCAN \cite{Ester1996}, as density based clustering technique. The presented approach performs a clustering on the search mask $\vecset{C}_s$ and uses the $\varepsilon$-neighborhood for an euclidean position and velocity distance between the cell $c$ and every other cell $c^*$. The resulting clusters $n$ are a set of cells 
\begin{equation}
	\begin{split}
	\vecset{C}_c^{(n)} = \Big\{ \vec{s}_c  \in \vecset{C}_s \; | \; &  \zweinorm{\vec{p}_c - \vec{p}_{c^*}} \leq \varepsilon_{p} \; \wedge \\ & \zweinorm{\vec{v}_c - \vec{v}_{c^*}} \leq \varepsilon_{v} \Big\},
	\end{split}
\end{equation} 
that represent potential detected objects. Firstly, using a maximum two dimensional position distance $\varepsilon_{p}$ ensures clustering of cells next to each other and secondly, the maximum velocity deviation $\varepsilon_{v}$ implicitly considers the cells orientation using the velocity in $x$- and $y$-direction. \figurename{\ref{subfig:c:gridmapextraction}} visualizes the DBSCAN clustering applied on the search mask, where cells with same color indicate a cluster.
\subsubsection{Validation Mask}
Obviously, since the search mask contains almost every occupied cell it is not possible to separate between background and other road users. Thus, a validation mask is used where each cell has to fulfill strict criteria. Additional, besides the cell state \eqref{eq:cellstategme}, for every cell a Mahalanobis distance 
\begin{equation}
d_0 = \sqrt{ 
	\vec{v}_c\trans
	\cdot
	\vec{P}_c^{-1}
	\cdot
	\vec{v}_c}
\end{equation}
relative to zero velocity is calculated \cite{Nuss2017}. A high value of $d_0$ indicates a great certainty that there is any movement. In addition, the absolute cell velocity~$v_c = \sqrt{v_x^2 + v_y^2}$ is calculated. The validation mask 
\begin{equation}
\begin{split}
\vecset{C}_v = \Big\{ \vec{s}_c  \in \vecset{C}_s \; | \; & p_c(O)  \geq \varepsilon_{p(O)} \;\wedge \; v_c  \geq  \varepsilon_{v_c} \; \wedge \;  \\
&  \sigma^2_{v_x v_x}  \leq \varepsilon_{\sigma^2_{v_x}} \; \wedge \; \sigma^2_{v_y v_y}  \leq \varepsilon_{\sigma^2_{v_y}} \; \wedge \; \\
&  d_0  \geq \varepsilon_{d_0}  \Big\}
\end{split}
\end{equation} 
examines the cells occupancy probability, absolute velocity and Mahalanobis distance to reach a minimum threshold. In addition, the velocity variances must not exceed a maximum threshold. \figurename{\ref{subfig:c:gridmapextraction}} visualizes the few remaining cells of the validation mask. 
\subsubsection{Validate Clusters}
Certainly, creating clusters on the search mask causes many false object hypotheses. Thus, the resulting clusters are validated, using cells of the validation mask. For each cluster, a minimum number of corresponding validated cells are required. For that reason, the ratio between the amount of cells $N_n$ in a cluster and the remaining number of validated cells $N_n^*$ is calculated. If the ratio 
\begin{equation}
	r_N  =  \dfrac{N_n^*}{N_n}
\end{equation}
exceeds a minimum threshold $\varepsilon_{r_N}$, the cluster is valid. Finally, the resulting clusters 
\begin{equation}
\vecset{C}_{c}^{(n')} = \Big\{ \vec{s}_c  \in \vecset{C}_c^{(n)} \; | \; r_N \geq \varepsilon_{r_N} \Big\},
\end{equation} 
are used to create objects. \figurename{\ref{subfig:e:gridmapextraction}} shows the validated and clustered cells.
\subsubsection{Object Creation and Label Assignment}
For each validated cluster $n'$ an object is created. Here, the cell states are used to estimate the objects position $\vec{p}_{rp} = [x,y]\trans$ with a corresponding reference point label~\eqref{eq:refpoint}, the mean absolute velocity $v$, mean orientation $ \varphi $, a bounding box $\vec{B}$ \eqref{eq:bb} as orientated rectangle of the corner points and an unique label~$\ell$. Additionally, in every time step, the last extracted grid map objects are predicted with a CV model to the current time. Afterwards, an association for the predicted and currently extracted objects is executed. This is performed, calculating an euclidean distance of the reference points and finding the optimal associations with the Hungarian Method \cite{Kuhn1955}. If two objects are successful associated the label $\ell$ of the predicted object is assigned to the currently extracted object. \figurename{\ref{subfig:f:gridmapextraction}} shows the resulting objects from the presented scene, including the corresponding cells transformed to the ego vehicle coordinates.  
Finally, the estimated object state
\begin{equation}
 	\hat{\vec{s}}_\mathcal{G} = \{\vec{p}_{rp}, v, \varphi, \vec{B}, \ell  \} \in \vecset{S}_\mathcal{G}
\end{equation}
is passed to the high-level fusion module as set of extracted grid map object $\vecset{S}_\mathcal{G}$ at every sample time $t_\mathcal{G}$.

\section{High-Level Fusion}
\label{sec:highlevelfusion}
In this section, the processing of the results of the functional modules in the perception layer is described. The main goal of the high-level fusion is to define a common representation for every object and fuse and validate the object estimates in a complete set of objects. 

\subsection{Fusion of Objects}
To receive a complete and accurate set of objects in the ego vehicles local environment, the tracks of the LMB filter and the extracted objects from the DOGMa are aggregated and fused to a common representation in the high-level fusion module. Here, the ego-stationary coordinates are used as base representation of all necessary information, e.g. object states or contextual information from the digital map. Since, the LMB filter and DOGMa are processed in vehicle coordinates, a simple transformation of the extracted objects to ego-stationary coordinates is performed. Hereafter, to simplify notations of the object states, $\hat{\vec{s}}_\mathcal{T}$ and $\hat{\vec{s}}_\mathcal{G}$ are in ego-stationary coordinates. Additionally, the digital map is transformed from UTM coordinates to ego-stationary coordinates. 
Furthermore, the LMB $\vecset{S}_{\mathcal{T}}(t_\mathcal{T}) $ and DOGMa $\vecset{S}_{\mathcal{G}}(t_\mathcal{G}) $ have different sample times, where each module transmits their result as soon the processing is finished. For this reason, the output of the modules is sequentially ordered in a waiting queue at the input of the high-level fusion module and processed in the correct order.

The high-level fusion aggregates every detected object of the functional modules and incorporates them in a complete set of objects $\vecset{S}_\mathcal{H}$. This set is sequentially updated with every sample from the functional modules. Because the state representation of the extracted grid map objects $\hat{\vec{s}}_\mathcal{G} $ slightly differs from the extracted tracks $\hat{\vec{s}}_\mathcal{T}$ a new meta object state
\begin{equation}
	\hat{\vec{s}}_\mathcal{H} = \{\vec{p}_{rp}, v, \varphi, \vec{B}, c, \ell, \eta, \hat{\vec{s}}_\mathcal{G}, t_\mathcal{G}, \hat{\vec{s}}_\mathcal{T}, t_\mathcal{T}  \}
\end{equation}
is created. Here, $\vec{p}_{rp} = [x,y]\trans$ is a two dimensional position at the reference point \eqref{eq:refpoint}, with an absolute velocity $v$ and an orientation $\varphi$. In addition, the length and width of the object is represented with the bounding box \eqref{eq:bb} and the object type $c$ is estimated. New meta objects attached with an unique label $\ell$ and comprise the corresponding time and state of the DOGMa and LMB, that created or updated the meta object. Since, in complex scenarios, a multi-object tracking or the DOGMa object extraction can generate false alarms or uncertain states a new parameter $\eta$, the confidence of an object is introduced. 
\subsection{Object Confidence}
To determine the quality of object estimates, received from either the DOGMa or LMB, a confidence value is calculated. For that reason, highly uncertain object estimates or false positives from the functional modules should be detected and not included in an update for the meta object. For this purpose, the confidence is determined using \textit{physical, module} and \textit{digital~map} constraints. 
\subsubsection{\textit{Physical} Constraint}
The \textit{physical} constraints validates the state $\hat{\vec{s}}_\mathcal{G} $ or $\hat{\vec{s}}_\mathcal{T} $ relative to a last meta object state $\hat{\vec{s}}_\mathcal{H}$. For that reason, physical possible motions of the objects are examined using the relative movement between the last and current sample. The validity is represented with the confidence $\eta(\hat{\vec{s}} | \mathcal{P})$.
\subsubsection{\textit{Module} Constraint}
For the \textit{module} constraint, the functional module specific characteristics are examined. Here, extracted tracks of the LMB filter estimate a covariance $\vec{P}$ of the object state and an existence probability $r$. These parameters are analyzed to confirm the objects state. Furthermore, objects of the DOGMa are checked, if they are detected over subsequent time steps. 

In addition, the confidence $\eta(\hat{\vec{s}} | \mathcal{E})$ of the functional modules $\mathcal{E = \{\mathcal{G,T} \}}$ increases, if an object is detected by both modules. This is evaluated in consideration of the modules specific sensors field of view. 
\subsubsection{\textit{Digital Map} Constraint}
The most significant weight generates the \textit{digital map} constraints. Here, the map data, described in Section \ref{seq:digitalmap}, is used to validate every objects state. First, the confidence  $\eta(\hat{\vec{s}} | \mathcal{M})$ reduces, when an objects position is clearly inside a building $\vecset{B}$. In addition, the position and orientation of the vehicles are validated using the lanes $\vecset{L}$ and corresponding rectangles $\vecset{R}$. Currently, the uncertainty of the global ego position is not considered due to the high precise DGPS localization system, but can be integrated when using a different approach. Finally, all constraints are evaluated to update an object. 
\subsection{Object Update}
To include a new object to the set of high-level fusion meta objects $\vecset{S}_\mathcal{H}$ the label of the DOGMa or LMB object is analyzed. If a label is unknown and the estimated confidence value
\begin{equation}
\eta = \eta(\hat{\vec{s}} | \mathcal{P}) \cdot  \eta(\hat{\vec{s}} | \mathcal{E}) \cdot   \eta(\hat{\vec{s}} | \mathcal{M})  \in \left(0,1 \right) 
\end{equation}
exceeds a minimum, a new meta object will be created using the corresponding object state and a new unique label. If the label $\ell_\mathcal{G}$ or $\ell_\mathcal{T}$ is already known, the object state is updated. Since, the high-level fusion receives sets of objects sequentially, an association between the already known objects from one module to the other module have to be identified. For that reason, the last sample is predicted with a simple CV model and a matrix of possible associations is created. Here, the optimal assignment is determined using the Hungarian method \cite{Kuhn1955} regarding an euclidean distance of the reference points. 

To update an object the confidence is examined. If $\eta$ exceeds a minimum, the position $\vec{p}_{rp}$, velocity $v$ and orientation $\varphi$ of the meta object is updated. Additionally, the length $l$ and width $w$ is updated regarding the reference point \eqref{eq:refpoint}. In particular, the following separations are considered:
\begin{equation}
	\begin{cases}
		 w \; &\text{when} \;  rp \in \{{\text{{b, f}}}\} \\
		 l \; &\text{when} \;  rp \in \{{\text{{l, r}}}\} \\
		 w, l \; &\text{when} \;  rp \in \{{\text{{bl, fl, fr, br}}}\}.
	\end{cases}
\end{equation}
Finally, the corresponding bounding box $\vec{B}$ is calculated using the updated width and length. If an object does not receive any samples over a short period, it will be removed from the set.

\section{Experiments}
\label{sec:experiments}
Evaluating the presented approach, the experimental vehicle of Ulm University \cite{Kunz2015} is used. Here, the vehicle is equipped with lidar, camera and radar sensors, which are used to record real world data in many different scenarios. The environment perception framework was evaluated in various scenes at urban and rural areas. To highlight the advantage of using a subsequent high-level fusion with a confidence value, two challenging scenarios are chosen as experimental results. 
The multi-object tracking with an LMB filter detects and tracks vehicles with a wide-angle camera and a frontal long-range radar. Here, extracted tracks with an existence probability $r > 0.2$ are passed to the high-level fusion. 
The DOGMa incorporates a frontal lidar sensor with an opening angle of $100^\circ $ and range of $100$m and a $360^\circ$ Velodyne PUK VLP-16 lidar mounted on the top of the vehicle with reflections up to $40$m. For the grid map object extraction the following thresholds are used:
${\varepsilon_{M_O} =  0.3} \text{, } {\varepsilon_{p(O)} = 0.8} \text{, } {\varepsilon_{v_c} =  0.3\frac{\text{m}}{\text{s}}}  \text{, } {\varepsilon_{\sigma^2_{v_x}} = 5} \text{, } \linebreak {\varepsilon_{\sigma^2_{v_y}} = 5} \text{, } {\varepsilon_{d_0} = 9} \text{, } {\varepsilon_{p} = 1.2\text{m}} \text{, } {\varepsilon_{v} = 1 \frac{\text{m}}{\text{s}}} \text{ and } {\varepsilon_{r_N} = 0.1} $. 
The DOGMa cell size is defined to $0.15 \text{m}  \times 0.15 \text{m}$ to achieve a total width $W = 120\text{m}$ and length $L = 120\text{m}$.

\begin{figure}[!t]
	\vspace{2mm}
	\centering  
	\includegraphics[width = 0.75\columnwidth, height=2cm]{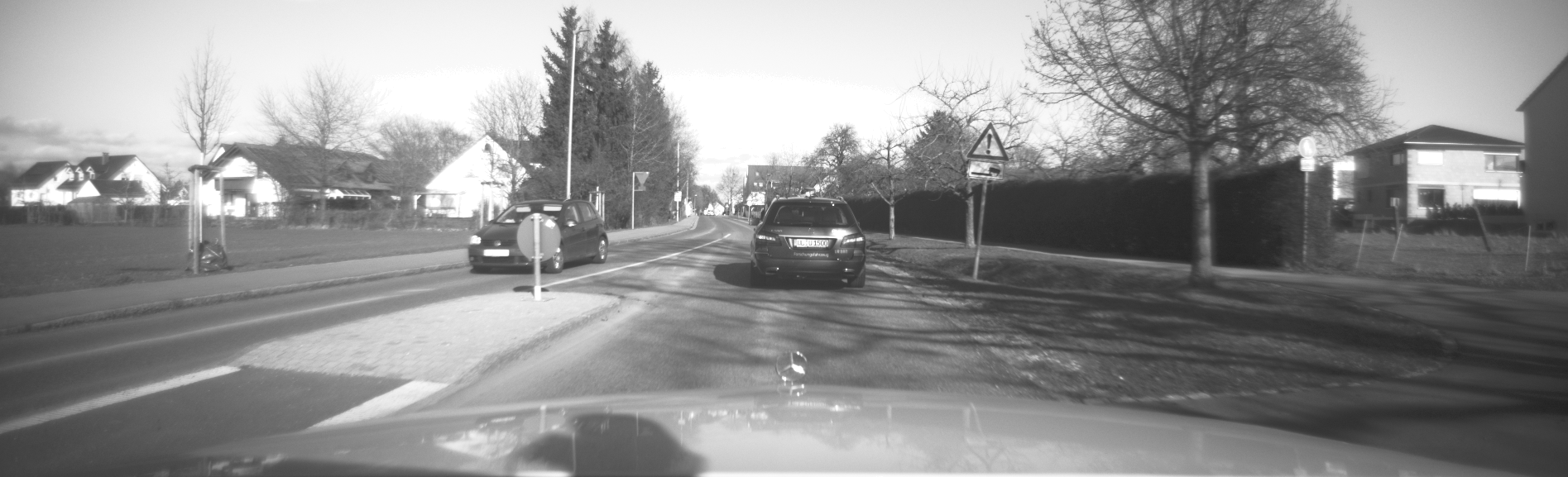}
	\begin{multicols}{2}
		\setlength{\figH}{5cm}
		\resizebox{1\columnwidth}{!}{
\small 
\input{img/tikz/lehr_gt_1032.tikz}} \\
		\setlength{\figH}{5cm}
		\resizebox{1\columnwidth}{!}{
\small 
\input{img/tikz/lehr_hlf_1032.tikz}} \\		
	\end{multicols}
	\vspace{-6mm}
	\caption{Exiting a roundabout, when a false track is generated with subsequent false radar measurements caused by a traffic island. On the top, camera image of the current scene. On the bottom left, extracted tracks (blue) and extracted grid map objects (red) with the ego vehicle (green) are shown. On the right the corresponding meta objects (magenta) are visualized. Additionally, the object (dashed black) with a low confidence value is not included in the set of meta objects.}
	\label{fig:lehr}
\end{figure}

\figurename{\ref{fig:lehr}} visualizes the first scenario. Here, the ego vehicle (green) is exiting a roundabout following a leading vehicle on a rural road. After exiting the roundabout, an oncoming vehicle is detected and false radar measurements are generated at a traffic island. Due to that, the LMB filter initializes a false track $\ell_\mathcal{T} = 7 $ (blue) and confirms it over subsequent frames. Here, the grid map objects (red) do not contain this object and consequently the \textit{module} constraint decreases. In addition, because of a high orientation error relative to the lane, the \textit{digital map} constraint further reduces the confidence value of the false track (dashed black). Finally, the false track is not validated and not included in the set of meta objects $\vecset{S}_\mathcal{H}$ (magenta). 

\begin{figure}[!t]
	\vspace{2mm}
	\centering  
	\includegraphics[width = 0.75\columnwidth, height=2cm]{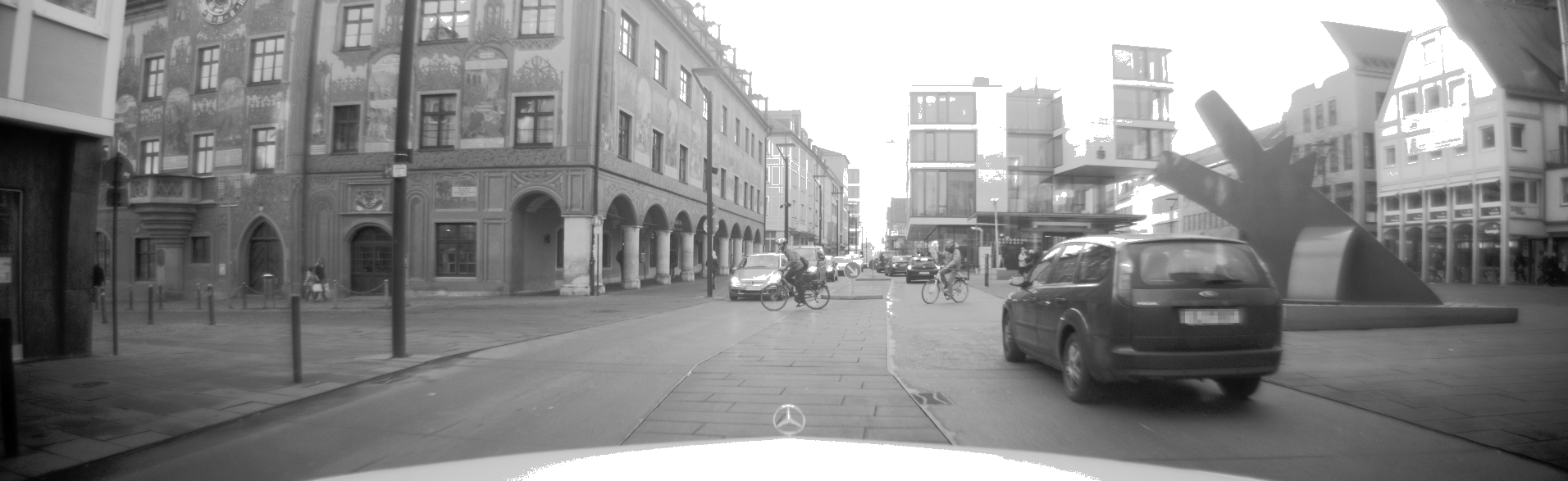}
	\begin{multicols}{2}
		\setlength{\figH}{6cm}
		\resizebox{1\columnwidth}{!}{
\small 
\input{img/tikz/mitte_gt_720.tikz}}
		\setlength{\figH}{6cm}
		\resizebox{1\columnwidth}{!}{
\small 
\input{img/tikz/mitte_gt_735.tikz}}
		\setlength{\figH}{6cm}
		\resizebox{1\columnwidth}{!}{
\small 
\input{img/tikz/mitte_gt_748.tikz}} \\
		\setlength{\figH}{6cm}
		\resizebox{1\columnwidth}{!}{
\small 
\input{img/tikz/mitte_hlf_720.tikz}}
		\setlength{\figH}{6cm}
		\resizebox{1\columnwidth}{!}{
\small 
\input{img/tikz/mitte_hlf_735.tikz}}
		\setlength{\figH}{6cm}
		\resizebox{1\columnwidth}{!}{
\small 
\input{img/tikz/mitte_hlf_748.tikz}} \\		
	\end{multicols}
	\vspace{-6mm}
	\caption{Complex inner city scenario with buildings (cyan), roads (gray) and the ego vehicle (green), showing a sequence of the presented approach. On the top, the first camera image of the sequence. On the left column, the extracted tracks (blue) and grid map objects (red) are shown subsequently from top to the bottom. On the right column, the corresponding meta objects (magenta) with not confident estimates (black dashed) in the last frame.}
	\label{fig:neuemitte}
\end{figure}

The second experiment is a complex inner city scenario, with a wide variety of different road users and stop-and-go traffic. Here, a lot of objects are occluded and correct associations between tracks and measurements are ambiguous. \figurename{\ref{fig:neuemitte}} shows the scene with its progressive frames. 
Here, the tracks (blue) and extracted grid map objects (red) are fused to the corresponding meta objects (magenta). 

First, a false grid map object with label $1103$ (red) is removed from the meta objects (magenta) because it is clearly inside a building. This objects was created due to lidar reflections from a buildings glass front, what can occur in common inner city scenarios. Additionally, two bicyclist are crossing the road in front of the ego vehicle. Consequently, the tracks with label $\ell_\mathcal{T} = 61 \text{ and } \ell_\mathcal{T} = 64$ (blue) are occluded and no measurements are generated. The LMB filters object estimation is now based on its prediction and the uncertainty increases, whereas the confidence value decreases in the last frame (dashed black). Finally, the environment perception comprises a complete set of meta objects including only confident object estimates.

\section{Conclusion}
\label{sec:conclusion}
In this work, an environment perception framework is introduced using various functional modules. For this reason, a common representation is developed using the ego-stationary coordinates to prevent uncertainties in the fusion of perceived objects. Furthermore, a multi-object tracking with an LMB filter and a DOGMa are implemented to detect and track road users in the local surroundings. Since, any object estimation method can generate uncertain states or false positives in complex scenarios a validation of the objects state is realized using a high-level fusion. To determine the quality of an object estimate a confidence value is introduced, that incorporates \textit{physical, module} and \textit{digital~map} constraints.

Obviously, the multi-object tracking and the DOGMa have strengths and weaknesses, that strongly depend on the current scenario, sensor setup and environment conditions. Especially, an adjustment of the parameters in the grid map object extracting algorithm has a strong impact on the performance. However, the main goal in the presented approach is, that a fusion of two independent functional modules for an object estimation results in a complete and accurate set of meta objects. Consequently, weaknesses of one functional module can be compensated utilizing the strength of the other module. 

For future work, a full evaluation on a public data set is necessary to show the generalization of the presented approach. In addition, the digital map will be extended with further context information for the object validation. 

\section*{Acknowledgment}
The research leading to these results was conducted within the Tech Center a-drive. Responsibility for the information and views set out in this publication lies entirely with the authors.


\bibliographystyle{IEEEtranBST/IEEEtran}
\bibliography{IEEEtranBST/IEEEabrv,bib/myBibTex}

\end{document}